\documentclass[sigconf, screen, authorversion, nonacm]{acmart}

\definecolor{minor}{HTML}{FF0000}
\usepackage{xcolor}

\usepackage{hyperref}
\usepackage[hyphenbreaks]{breakurl}
\usepackage{graphicx}
\usepackage{csquotes}
\usepackage{makecell}
\usepackage{subcaption}
\usepackage{svg}
\usepackage{enumitem}

\AtBeginDocument{%
  \providecommand\BibTeX{{%
    \normalfont B\kern-0.5em{\scshape i\kern-0.25em b}\kern-0.8em\TeX}}}
 
\setcopyright{acmcopyright}
\copyrightyear{2018}
\acmYear{2018}
\acmDOI{10.1145/1122445.1122456}

\acmConference[Woodstock '18]{Woodstock '18: ACM Symposium on Neural
  Gaze Detection}{June 03--05, 2018}{Woodstock, NY}
\acmBooktitle{Woodstock '18: ACM Symposium on Neural Gaze Detection,
  June 03--05, 2018, Woodstock, NY}
\acmPrice{15.00}
\acmISBN{978-1-4503-XXXX-X/18/06}

\begin{document}

\title{The use of deception in dementia-care robots: Should robots tell ``white lies'' to limit emotional distress?}


\author{Samuel Rhys Cox}
\email{samcox@comp.nus.edu.sg}
\orcid{0000-0002-4558-6610}
\affiliation{%
  \institution{National University of Singapore}
  \country{}
  }

\author{Grace Cheong}
\email{llcheong@smu.edu.sg}
\orcid{0000-0002-6128-5834}
\affiliation{\institution{Centre for Research on Successful Ageing (ROSA),\\ Singapore Management University}
\country{}
  }

\author{Wei Tsang Ooi}
\email{ooiwt@comp.nus.edu.sg}
\orcid{0000-0001-8994-1736}
\affiliation{\institution{National University of Singapore}
\country{}
  }

\begin{abstract}
With projections of ageing populations and increasing rates of dementia, there is need for professional caregivers.
Assistive robots have been proposed as a solution to this, as they can assist people both physically and socially.
However, caregivers often need to use acts of deception (such as misdirection or white lies) in order to ensure necessary care is provided while limiting negative impacts on the cared-for such as emotional distress or loss of dignity.
We discuss such use of deception, and contextualise their use within robotics. 

\end{abstract}



\maketitle


\section{Introduction}
Population ageing coupled with lower fertility rates is an increasing concern for many countries and with a growing population of older adults, there is an increasing demand for professional caregivers to address the care needs of older adults at a societal level. This is particularly so for persons with dementia. Research estimates that the number of people living with dementia will approximately triple from 57 million in 2019 to 153 million by 2050 \cite{AlzheimerResearchUK}. Cognitive functioning deteriorates as the disease progresses, inhibiting the independence and functioning of persons afflicted with the disease. Depending on the severity of impairment, persons with dementia may require extensive, consistent care in daily living. 

To address this gap, research has pointed towards the use of robots to address the care-giving needs of older persons \cite{bodenhagen2019robot,feil2011socially}. Assistive robots have been deployed in a multiplicity of care settings \cite{martinez2021assistive,asgharian2022review,ghafurian2021social,khaksar2023robotics}. 
However, unlike human care-givers, assistive robots at present lack the reflexivity to respond according to the changing needs and behaviour of their cared-fors. This then also raises the question of how and if robots should adapt their response according to the changing cognitive state of their cared-fors. 

On from this, advice for human carers would adapt depending on the severity of the cognitive decline associated with the cared-for's dementia. Strategies in early stages of dementia emphasise orienting the recipient's consciousness to be more reality based \cite{carrion2013cognitive,kanov2017sorry} (such as via stating the year, time of day and current weather, and the cared-for's name, age and significant relationships). However, in later stages of dementia such an orientation towards reality may cause emotional distress to the cared-for \cite{james2003lying} and different approaches may be needed \cite{elvish2010lying,erdmann2016conditions}.
For this reason, advice would dictate that the care-giver could deploy acts that may be considered deceptive in order to fulfil the care needs of cared-for while limiting emotional distress and loss of dignity \cite{oye2020informal,cantone2019lying}. In addition, while it is in the interests of the care-giver to ensure that the cared-for is given sufficient care (such as ensuring hygiene is maintained, and that cared-for are safe) delusions from cognitive decline may conflict with these needs and cause distress. These equally would lead to care-givers potentially using techniques that are in some way acts of deception.

A robot that fully emulates a human carer would then, on occasion, deceive those that it is caring for.
Related to this, schools of thought for robotic deception could be seen in two camps: those who oppose the use of deception as \cite{sharkey2021we} (perhaps due to leading to over-dependence, loss of human relationships or lost sense of reality); and those who view deception as a necessary and inevitable act \cite{carli2023reconsidering,isaac2017robots} (as robots continue to anthropomorphise and adopt human characteristics, thereby increasing acceptability and effectiveness \cite{short2010no,esposito2018engagement}).
At a philosophical and ethical level, the context of robot lies have been discussed within healthcare. For example, Matthias described that lies should always be in the patient’s best interests, increase patient autonomy, not lead to harm, and be transparent in deception \cite{matthias2015robot},
yet these do not address precise strategies that could be used in dementia-care by assistive robots, or people's perception of these.
On from this, we discuss a number of strategies (involving deception) for dementia-care that are either recommended practice, or have been reported as used by care-givers.
Afterwards, we discuss potential implications and concerns that could be raised if these same strategies were used by assistive robots.
For example, when could a robot need to deceive people in order to better complete its tasks, and why would it need to deceive someone rather than revealing the truth?

\section{Human carers and uses of deception}
\label{sec:deception}

A goal of human carers is to deliver necessary care, while balancing emotional distress, dignity, and humanity in order to limit both physical and emotional harm.
For example, while allowing for autonomy and maintaining independent activities improves well-being and cognitive functions, if such freedom exists, so too does the possibility for resistiveness due to conflicting wants between care-givers and the cared-fors \cite{watts2019speaking}, with such resistiveness causing distress to both parties \cite{mortensen2022strategies}. Additionally, in the case of persons with dementia in particular, cognitive impairments may hinder the ability to make rational decisions and complete tasks that are beneficial to one's well-being. 

As a consequence of this, human carers may use techniques to lessen resistiveness to care that could be seen as deceptive, such as  therapeutic lying \cite{cantone2019lying,elvish2010lying,james2003lying,james2006lying}, informal use of restraint \cite{oye2020informal}, or forms of indirect coercion \cite{sander2005concealment}.
While some of these techniques are widely adopted (if not sometimes contentiously \cite{oye2020informal,caiazza_inbook,caiazza2016untruths,cantone2019lying}) among carer-givers, additional debate is needed for their potential use by assistive robots.

Specifically, therapeutic lying (lying that is used for the benefit of the cared-for, rather than for the care-giver) can be used to limit emotional distress while delivering care \cite{cantone2019lying,elvish2010lying,james2003lying,james2006lying}.
Therapeutic lies could be used for ``tricks'' \cite{elvish2010lying} such as to simplify ingestion of medication \cite{cantone2019lying}, to avoid aggressive behaviour, to limit time spent giving explanations, to go along with a care-for's misconception, and to alleviate stress (see \cite{cantone2019lying} for examples of therapeutic lies). For example, a person with dementia may ask their care-giver where their deceased parent is, to which the care-giver could reply ``they'll come tomorrow'' \cite{cantone2019lying}. 
These therapeutic lies involve some form of deception based in verbal communication, that while uncertain in their levels of ethics and acceptability, would become more feasible as the abilities of language models improves.

Similarly, {\O}ye and Jacobsen surveyed nursing homes in Norway, and identified five distinct types of ``informal restraint'' enacted by caregivers \cite{oye2020informal}. While not acts of formal restraint (such as physical restraint that would limit autonomy and cause physical and psychological pain), informal restraint limits freedom of movement and personal preference in order to provide care. Specifically, they identified diverting residents’ attention (such as showing photographs to distract a person with dementia during washing that they may otherwise protest to); white lies (to affirm the perceived realities of persons with dementia); persuasion and interpersonal pressure; offers (to incentivise adherence to the care-giver's requests); and threats (such as sending them back to their room to enforce compliance) as forms of informal restraint.
They additionally identified ``grey-zone'' constraints, such as seating a resident in a chair that is deep and low so that they cannot get up on their own, and therefore reduce their chance of wandering and hurting themselves.


\section{Discussion}

We have provided an overview of some deception-based techniques that care-givers may use when providing care for people with dementia.
We will now discuss issues related to their potential use and application by assistive robots.

While the use of deception techniques such as therapeutic lying is well investigated for both its prevalence and acceptability in care-giving environments \cite{cantone2019lying}, such beliefs are not well investigated with regards to assistive robots, and there are potential ethical concerns that a person with dementia may not be able to distinguish between a relationship with a robot and a relationship with a person \cite{saetra2020foundations}.
However, previous literature has demonstrated that perhaps there is a difference in attitude between the philosophical ideals and clinical needs and practice when deploying forms of deception in dementia-care.
For example, Koh et al. \cite{koh2022determinants} investigated the use of robot pets in nursing homes for dementia, and found that, while people were apprehensive about the potential for deception (i.e., people with dementia believing a robot pet to be real), once stakeholder's personally experienced real-world use of robots, most were comfortable with its adoption.
Although it is difficult to expand such a discussion to the more direct forms of deception in Section \ref{sec:deception}, it could be forseen that such acceptance would be more normalised and likely as adoption and capabilities of robotics increases.

In addition, it is important to note that some of the uses of deception may be related to lack of resources for care-givers \cite{oye2020informal,cantone2019lying}. 
For example, the use of grey-zone restraints \cite{oye2020informal} could be avoided in a well-resourced home, or if a vision of a future with more carer-givers due to assistive robots is met. 
Such forms of deception are less ethically and morally defensible and (while ensuring the safety of the cared-for) would perhaps be less desirable and less acceptable if used by a robot.

The changing nature of interventions used by care-givers could also be challenging for robotic assistants to overcome. For example, a therapeutic lie may have limited effectiveness the more it is used, and additional alternative therapeutic lies may need to be used in order to still deliver care while limiting the emotional distress of the cared-for \cite{cantone2019lying}. 
It is unclear whether giving a robot the freedom and flexibility to devise such deceptions would be seen as acceptable and ethical, and if this necessary flexibility would have the potential for harm.
This changing nature could also lead to technical challenges due to differences in each person and their perceived independent autonomy and application of care as such. 
For example, how would such a robot be designed to differentiate and perceive the level of cognitive abilities of individuals and thereby the potential application of deception to aid in delivery of care?
Such a robot would need to be able to distinguish between situations where it can apply deception (and to whom) and situations and persons where this is not appropriate, or where orientating in reality (rather than using therapeutic lies for example) would be acceptable.


In addition, there are also potential legal issues caused by deception-led robotic interventions with questions surrounding liability if such methods are seen as  as harmful, distressing and coercive \cite{loh2018medicine,gunkel2022should}.
Furthermore, if such interactions are (presumably) recorded and actionable to one robotics corporation this centralised control could be more liable to litigation, or in need of more strict oversight.

\section{Conclusion}
In conclusion, we have discussed a number of techniques (that adopt forms of deception) used by human care-givers when caring for people with dementia. 
While  we cannot provide a clear consensus on the acceptability, effectiveness or ethics of adopting each technique, we would advise that assistive robots behave so as to enhance emotional and physical well-being \cite{coeckelbergh2011artificial}.
With robots and language models becoming more capable and prevalent in the provision of care, it is hoped that our discussion will lead researchers to reflect on the likely potential adoption of such deceptive techniques, as well as draw attention to the need for additional studies of robotic ethics in innovative contexts of use \cite{stahl2016ethics}.

\begin{acks}
This research is part of the programme DesCartes and is supported by the National Research Foundation, Prime Minister’s Office, Singapore under its Campus for Research Excellence and Technological Enterprise (CREATE) programme.
\end{acks}
\bibliographystyle{ACM-Reference-Format}
\bibliography{sample-base}

\end{document}